\documentclass[conference]{IEEEtran}
\IEEEoverridecommandlockouts    

\usepackage{multirow}
\usepackage{lipsum}
\usepackage{amsmath}
\usepackage{amssymb}
\usepackage[ruled,vlined]{algorithm2e}
\usepackage{graphicx}
\graphicspath{{./Figures/}}

\title{\LARGE \bf High-Slip-Ratio Control for Peak Tire–Road Friction Estimation Using Automated Vehicles}

\begin{document}
\IEEEoverridecommandlockouts
\setlength{\textheight}{9.25in}

\author{Zhaohui Liang, Hang Zhou, Heye Huang, and Xiaopeng Li
\thanks{Authors are with the Civil \& Environmental Engineering Department, University of Wisconsin, Madison, Madison, WI({\tt\small \{zhaohui.liang, hzhou364, hhuang468,xli2485\}@wisc.edu})}%
}
	
	\maketitle
	\thispagestyle{empty}
	\pagestyle{empty}
	\begin{abstract}
		Accurate estimation of the tire-road friction coefficient (TRFC) is critical for ensuring safe vehicle control, especially under adverse road conditions. However, most existing methods rely on naturalistic driving data from regular vehicles, which typically operate under mild acceleration and braking. As a result, the data provide insufficient slip excitation and offer limited observability of the peak TRFC. This paper presents a high-slip-ratio control framework that enables automated vehicles (AVs) to actively excite the peak friction region during empty-haul operations while maintaining operational safety. A simplified Magic Formula tire model is adopted to represent nonlinear slip–force dynamics and is locally fitted using repeated high-slip measurements. To support safe execution in car-following scenarios, we formulate a constrained optimal control strategy that balances slip excitation, trajectory tracking, and collision avoidance. In parallel, a binning-based statistical projection method is introduced to robustly estimate peak TRFC under noise and local sparsity. The framework is validated through both closed-loop simulations and real-vehicle experiments, demonstrating its accuracy, safety, and feasibility for scalable, cost-effective roadway friction screening.
	\end{abstract}
	
	\section{Introduction}
	\label{sec:introduction}
	In adverse weather conditions such as rain, snow, or ice, insufficient tire–road adhesion is a major contributor to vehicle skidding, extended braking distances, and loss of control, significantly increasing crash risk~\cite{wallman2001friction, huang2022risk}. Accurate estimation of the tire–road friction coefficient (TRFC) is therefore critical for enhancing traffic safety and supporting infrastructure asset management~\cite{elkhazindar2022incorporating}. As a core parameter in both longitudinal and lateral vehicle dynamics, the peak TRFC defines the maximum traction available and plays a decisive role in acceleration, braking, and stability control~\cite{li2024capturing, hu2020real}.

Current TRFC estimation approaches can be broadly categorized into two groups: (1) skidding-based estimation, and (2) vehicle-based estimation using onboard sensors. The skidding-based estimation employs mechanical friction testers, such as the Locked-Wheel Skid Trailer (LWST), Mu-Meter, and Sideway-force Coefficient Routine Investigation Machine (SCRIM), are commonly used by road agencies to quantify pavement friction~\cite{mccarthy2021comparison, kumar2023state}. These devices induce slip through controlled mechanisms to directly measure tire–road interaction. While highly accurate in controlled settings, their deployment is limited by cost, operational complexity, and inability to provide continuous or geometry-flexible measurements (e.g., on ramps or curves)~\cite{hall2009guide, quiros2018interconversion}. For instance, the Locked-Wheel Skid Tester (LWST) measures friction under full-sliding conditions using a locked test wheel, which limits both its operating speed and its applicability on diverse roadway geometries, and may restrict its ability to capture the true peak friction value~\cite{de2019locked,smadi2025impact}. The SCRIM device, by contrast, employs a yawed (side-force) test wheel that enables continuous friction measurements at traffic speeds; however, it requires regular calibration and its Sideway-Force Coefficient (SFC) measurements can be affected by dynamic variations in vertical load, particularly on bends and geometrically complex road sections~\cite{roe2005recent}.

Vehicle-based estimation methods infer TRFC indirectly from onboard sensor data, such as wheel speeds, inertial measurements (IMUs), global positioning system (GPS), and brake signals. These methods offer greater scalability than mechanical testers and are compatible with modern connected or autonomous vehicle platforms. They can be broadly divided into regression-based and learning-based approaches.

\textbf{Regression-based methods} rely on simplified physical models of vehicle dynamics (e.g., 1-DOF, bicycle, or 3-DOF models) combined with empirical or semi-empirical tire models, such as the Magic Formula, brush model, or Burckhardt model~\cite{gustafsson1997slip, singh2015estimation, wang2022tire}. These models typically use measured accelerations, forces, and slip ratios to estimate friction. However, under naturalistic driving, vehicles operate at low slip ratios (e.g., $\kappa < 0.05$), which are insufficient to capture the nonlinear regime near peak TRFC. Furthermore, many such models assume constant vertical loads and neglect dynamic load transfer during aggressive maneuvers~\cite{alvarez2005dynamic, matuvsko2008neural}, leading to biased or conservative estimates.

\textbf{Learning-based methods} extend beyond the assumptions of physical models by training machine learning algorithms, such as neural networks, support vector machines, time series analysis \cite{cao2023vehicle} and deep architectures (CNNs~\cite{tian2024recent}, LSTMs~\cite{xiao2022novel}, and Transformers~\cite{ghorbani2025robust}), to map sensor data directly to TRFC estimates. Some also incorporate visual features from onboard cameras to infer road texture and surface condition~\cite{zhao2024road}. While these models can capture complex nonlinearities, their performance heavily depends on the diversity of the training data. Critically, the scarcity of high-slip samples in naturalistic datasets limits their ability to generalize to peak TRFC conditions. Moreover, the lack of physical interpretability raises concerns in safety-critical applications.

To enhance robustness, many TRFC estimation methods use multi-sensor fusion or observer-based approaches, often combining wheel-speed, inertial, and brake information through Kalman filtering techniques (e.g., EKF, UKF, CKF) \cite{canudas1999observers, mu2003estimation}, and some include brake-pressure excitation for surface classification ~\cite{yiugit2023estimation}. While effective in noise reduction, these approaches remain passive and generally lack the high-slip excitation required to reveal peak TRFC, yielding mostly lower-bound estimates.

\subsection{Proposed Approach and Contributions}
To address these limitations, we propose a novel high-slip-ratio control and estimation framework that actively excites tire–road interaction using automated vehicles (AVs) during empty-haul operations. This approach leverages the operational flexibility of AVs to perform controlled acceleration and deceleration maneuvers, enabling safe and repeatable access to high-slip regimes—even under worst-case assumptions. Such conditions are difficult to achieve using conventional regular vehicles (RVs) constrained by safety and comfort considerations.

The proposed framework incorporates a computationally efficient tire model derived from a Magic Formula, justified through sensitivity analysis. To enhance robustness against sensor noise and local variability, a binning-based statistical projection method is employed to estimate peak TRFC from high-slip data. A constrained optimal control strategy is formulated to ensure safe slip excitation while satisfying car-following safety constraints. The effectiveness of the overall framework is validated through both closed-loop simulations and real-world field experiments.

The key contributions of this work include:
\begin{itemize}
    \item An active high-slip-ratio excitation strategy using AVs during empty-haul trips to enable accurate peak TRFC estimation.
    \item A constrained optimal control design that balances slip excitation, safety, and trajectory tracking in car-following scenarios.
    \item End-to-end validation through simulation and on-road field testing, demonstrating the method's accuracy, robustness, and scalability.
\end{itemize}

The rest of the paper is organized as follows. Section \ref{sec:method} introduces the proposed high-slip-ratio control strategy and TRFC estimation framework. Section \ref{sec:exp} presents simulation and experimental results validating the effectiveness of the method. Finally, Section \ref{sec:conclusion} concludes the paper and discusses future research directions.
	
	\section{Methodology}
    \label{sec:method}
	
Accurate estimation of the peak TRFC requires active excitation of the high-slip regime, which is typically unreachable under normal driving. This section introduces a simplified tire model, a statistical projection-based estimator, and a high-slip-ratio control framework designed for AVs. The proposed method ensures interpretable TRFC estimation while maintaining operational safety in realistic car-following scenarios.

In the context of longitudinal vehicle dynamics, the TRFC exhibits a nonlinear relationship with the slip ratio. At low slip ratios (typically below 0.1), the TRFC increases rapidly and reaches a peak at the critical slip ratio (CSR). Beyond this point, it gradually decreases and stabilizes at a lower, quasi-constant level. The peak TRFC can vary significantly depending on tire and pavement characteristics, such as tire type and surface conditions. However, under normal driving conditions, vehicles seldom generate sufficient slip to reach this peak. Most operations maintain slip ratios below 0.05~\cite{gerard2006tire}, limiting access to the high-friction regime and making it difficult to observe or learn the full TRFC profile. Meanwhile, direct measurement using dedicated equipment such as skid testers or instrumented trailers remains costly and operationally impractical on a large scale. These constraints present a fundamental challenge to both model-based and learning-based friction estimation methods, which often lack accurate peak friction data for training or calibration. To address this gap, we propose a slip-ratio control strategy that deliberately induces high-slip conditions during safe vehicle maneuvers. Furthermore, we mathematically prove that, assuming a fixed CSR, increasing the instantaneous acceleration reduces the prediction error of peak TRFC. This approach enables reliable access to high-friction data and supports interpretable, data-driven friction estimation under real-world constraints.

\subsection{Definition of Slip Ratio}

The slip ratio \( \kappa \) quantifies the relative longitudinal motion between a vehicle’s tire and the road surface, serving as a key variable in characterizing tire force generation. Its definition varies depending on whether the vehicle is accelerating or braking. However, to ensure consistency and applicability across different driving modes, we adopt the following unified formulation:
\begin{align}
    \kappa = \frac{V_{\text{wheel}} - V_{\text{vehicle}}}{\max(V_{\text{vehicle}}, \epsilon)}
    \label{eq:slip_ratio}
\end{align}
where \( V_{\text{wheel}} = \omega \cdot R_e \) is the tangential velocity at the tire circumference, computed from the wheel angular velocity \( \omega \) and the effective rolling radius \( R_e \). \( V_{\text{vehicle}} \) denotes the vehicle’s longitudinal velocity measured at the center of gravity. A small positive constant \( \epsilon \) is included in the denominator to prevent division by zero and ensure numerical stability during low-speed or standstill conditions. This normalized definition allows for seamless integration into the tire force model in Section \ref{sec:magic_tire} and the slip-ratio-based control strategy introduced later in Section \ref{sec:high_slip_controller}. Notably, positive slip ratios (\( \kappa > 0 \)) typically correspond to driving or acceleration scenarios, while negative values reflect braking or wheel lockup conditions.

\subsection{Vehicle Dynamic Model}

\begin{figure}
    \centering
    \includegraphics[width=0.8\linewidth]{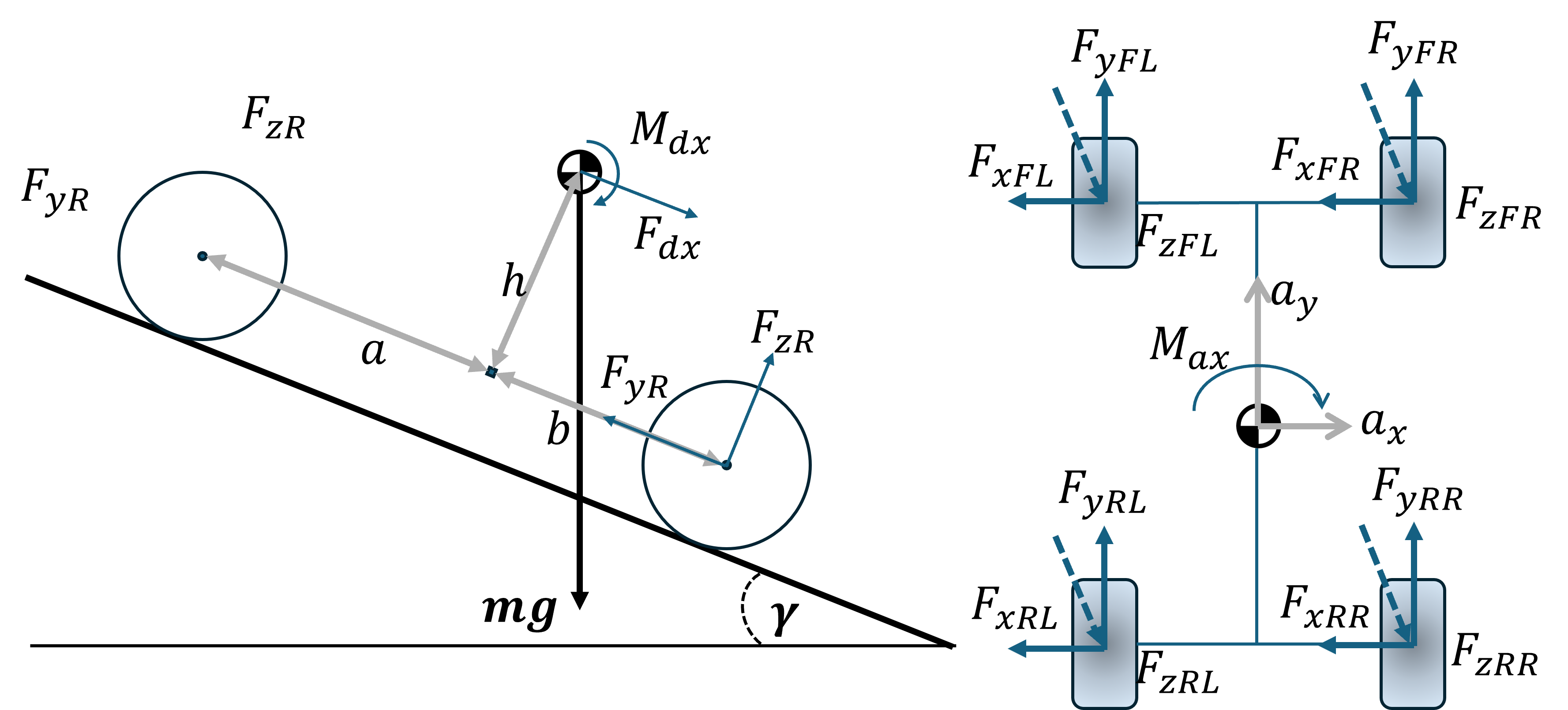}
   \caption{Four-wheel Vehicle model considering slope}
    \label{fig:vehicle model}
\end{figure}

To account for variations in longitudinal tire forces due to vertical load transfer, slope, and aerodynamic drag, we model the vehicle as a four-wheel system, as illustrated in Figure~\ref{fig:vehicle model}. The longitudinal dynamics follow Newton's second law:
\begin{align}
    m a_y = \sum_{i \in \{\text{F,R}\},j \in \{\text{L,R}\}} F_{yij} - F_{dy} - mg \sin(\gamma),
\end{align}
where $m$ is the vehicle mass, $a_y$ is the longitudinal acceleration of the center of gravity (CG), $F_{yij}$ denotes the longitudinal tire force at wheel $ij$ ($i \in \{\text{F,R}\}$ for front/rear and $j \in \{\text{L,R}\}$ for left/right), $F_{dy} = \frac{1}{2}\rho C_d A_f v^2$ is the aerodynamic drag, and $\gamma$ is the road slope angle.

The dynamic normal load on each axle is affected by pitch motion and CG height:
\begin{align}
    F_{zF} & = \frac{mg}{2L}(b - h\frac{a_y}{g}), \\
    F_{zR} & = \frac{mg}{2L}(a + h\frac{a_y}{g}),
\end{align}
where $a$ and $b$ are the longitudinal distances from CG to the front and rear axles, respectively, $L = a + b$ is the wheelbase, and $h$ is the CG height. $F_{zF}$ and $F_{zR}$ denote the total normal loads on the front and rear axles. This model enables accurate estimation of slip ratios and tire forces under dynamic conditions. Each longitudinal tire force $F_{yij}$ is computed as $F_{yij} = \mu_{ij} F_{zij}$, where the friction coefficient $\mu_{ij}$ depends on the slip ratio $\kappa_{ij}$.

\subsection{Magic Formula}
\label{sec:magic_tire}

The Magic Formula is a widely adopted empirical model \cite{pacejka1992magic} for capturing the nonlinear relationship between tire forces and slip, both in the longitudinal and lateral directions. Due to its excellent performance across a wide range of conditions, it is commonly referred to as the “Magic Formula.” In this study, we focus exclusively on modeling the longitudinal tire force, which is expressed as:
\begin{align}
    F = D \cdot \sin\left( C \cdot \arctan\left( Bx - E \cdot (Bx - \arctan(Bx)) \right) \right)
\end{align}
where $F$ is the longitudinal force, $x$ is the slip ratio, and $B$, $C$, $D$, $E$ are empirical parameters used to fit the model to experimental data. The parameter $E$, often referred to as the curvature factor, influences the shape of the force curve beyond the peak.

\begin{figure}[ht]
    \centering
    \includegraphics[width=1\linewidth]{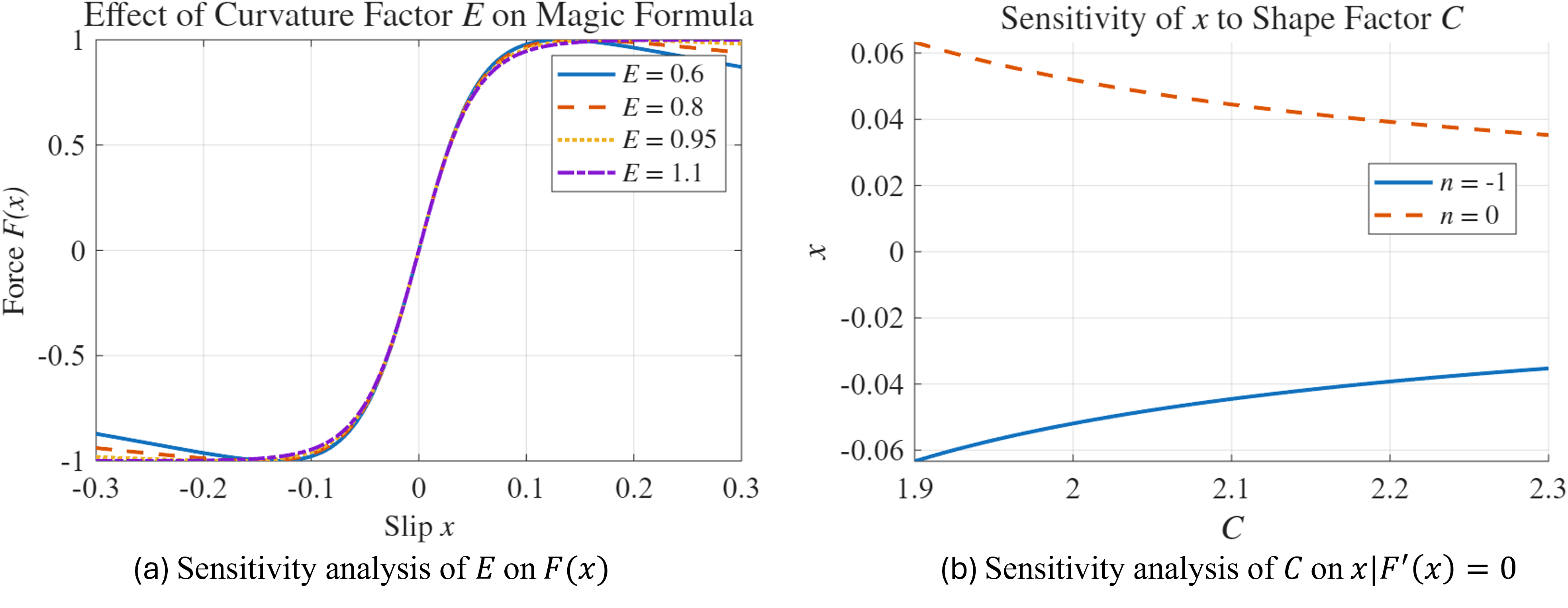}
    \caption{Sensitivity analysis of x to shape factor C.}
    \label{fig:EC vary}
\end{figure}

To examine the sensitivity of $E$, we fix $B$, $C$, and $D$ and vary $E$. As shown in Figure~\ref{fig:EC vary}(a), the curvature factor has little influence on the peak tire–road friction coefficient (TRFC) or the critical slip ratio (CSR). Therefore, to balance computational efficiency and model fidelity during real-time estimation of peak TRFC, we fix $E = 1$. The simplified Magic Formula then becomes:
\begin{align}
    F(x) = D \cdot \sin\left( C \cdot \arctan\left( \arctan(Bx) \right) \right)
    \label{eq:simplified Magic}
\end{align}

Taking the derivative of Equation (\ref{eq:simplified Magic}) with respect to $x$, we obtain:
\begin{equation}
\begin{aligned}
    F'(x) 
    &= D \cdot C \cdot \cos\left(C \cdot \arctan(\arctan(Bx))\right) \\ 
    &\cdot \left( \frac{B}{\left(1 + (\arctan(Bx))^2\right)(1 + (Bx)^2)} \right)
\end{aligned}
\end{equation}

To identify the CSR, we solve $F'(x) = 0$. Since the second term in the product is always positive, we set the cosine term to zero:
\begin{align}
    &\arctan(\arctan(Bx)) = \frac{\pi}{2C} + \frac{n\pi}{C}, \quad n \in \mathbb{Z} \\
    &x = \frac{1}{B} \tan\left( \tan\left( \frac{\pi}{2C} + \frac{n\pi}{C} \right) \right), \quad n \in \mathbb{Z}
\end{align}

Empirical studies suggest that the shape factor $C$ typically lies in the range $[1.9, 2.3]$. To analyze the sensitivity of CSR to variations in $C$, we fix $B$ and evaluate $x$ under the conditions $n = 0$ and $n = 1$, which correspond to the first two critical points due to the symmetry of the Magic Formula. As shown in Figure~\ref{fig:EC vary}(b), the variation of $C$ has limited influence on CSR. Thus, for practical purposes, we can fix the CSR by selecting an appropriate value of $B$.

\subsection{Error measurement of prediction}
\label{sec: Error measurement}

For a discrete data collection system, and in the absence of prior knowledge of the simplified Magic Formula tire model, we estimate the maximum tire–road friction coefficient (TRFC) independently at each sampled time point. Assuming a fixed critical slip ratio (CSR) by selecting a constant parameter $B$, we locally fit the simplified Magic Formula using observed data and nonlinear optimization.

Specifically, we apply a bounded optimization method (e.g., MATLAB’s \texttt{fmincon}) to identify the best-fitting parameters $C$ and $D$ at each time step by minimizing the difference between the measured longitudinal force and the model prediction. This point-wise fitting process yields a set of local TRFC estimates. 

\begin{figure}
    \centering
    \includegraphics[width=0.7\linewidth]{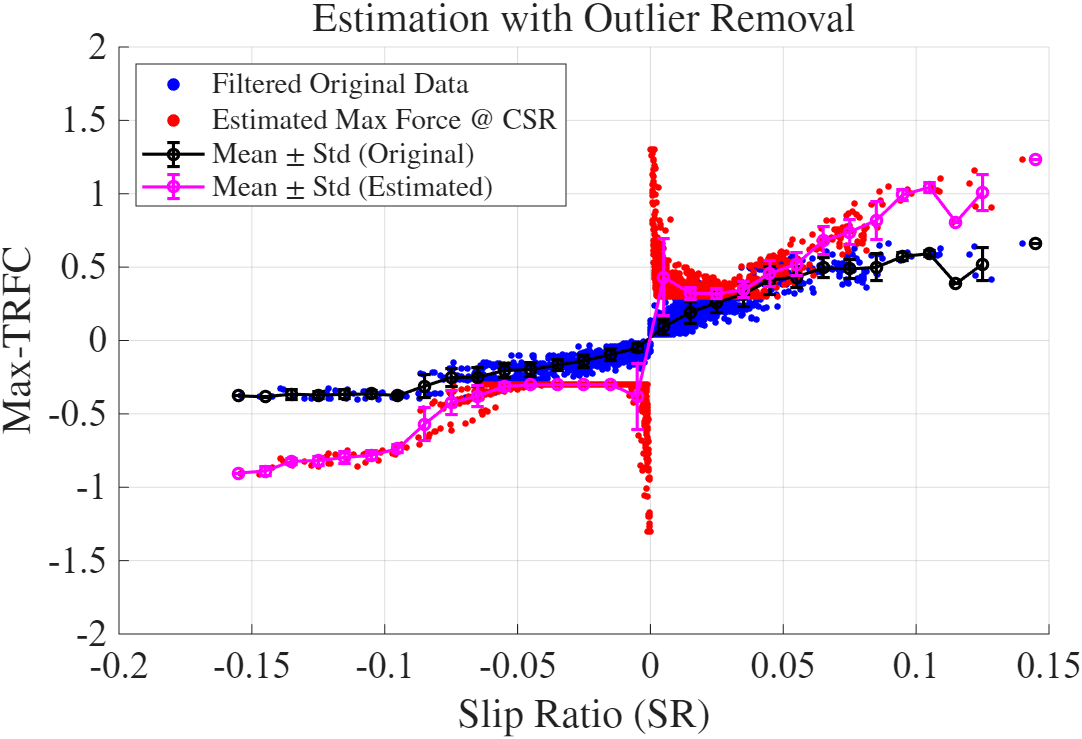}
    \caption{Illustration of Estimation Error Quantification}
    \label{fig:Error measurement}
\end{figure}

Because the optimization is performed independently at each point and relies on limited local measurements, the resulting Max-TRFC estimates may exhibit variability due to sensor noise, data sparsity, and local fitting uncertainty, as illustrated in Figure \ref{fig:Error measurement}. To quantify this variability, we define an error term as follows:

\begin{enumerate}
    \item Aggregate the data based on slip ratio for error analysis. Define a set of bins $\mathcal{I}$ with a bin width of 0.01, covering the range from 0.01 to 0.3.
    \item For each bin $i\in \mathcal{I}$, define the set of predicted peak TRFC value as  $y_i:=\{y_{ij}\}_{j\in\mathcal{J}_i}$ where each $y_{ij}$ is an individual prediction within bin $i$.
    \item  Let $y:=U_{i\in\mathcal{I}}y_i$ denote the set of all predicted peak TRFC value across all bins.
    \item Let $\bar{y}$ denote the true or referenced peak TRFC value.
    \item The sample mean of predicted peak TRFC value in bin $i$ is defined as: $\hat{y}_i:= \frac{1}{|\mathcal{J}_i|}\sum_{j\in\mathcal{J}}y_{ij}$ 
    \item Let $\delta_i:=STD(\hat{y}_i)$ denote the standard deviation of prediction in bin $i$ and $\bar{y}_i:=\mathbb{E}(\hat{y}_i)$ be the empirical expectation.
    \item Define the mean squared error for bin $i$ with respect to the true peak TRFC as $e_i = \mathbb{E}(y_{ij}-\bar{y})^2$
\end{enumerate}

Using the bias-variance decomposition, the mean squared error term $e_i$ can be rewritten as: 
\begin{align}
    e_i=\frac{1}{|\mathcal{J}_i|}\sum_{j\in\mathcal{J}_i}(y_{ij}-\hat{y_i})^2 + (\hat{y}_i-\bar{y})^2
    \label{MSE}
\end{align}

Using Equation \ref{MSE}, we establish the relationship between estimation error, CSR, and the peak TRFC. Physically, the peak TRFC can be considered equivalent to the feasible maximum acceleration. Therefore, we define $e(a_t)$ as a function that maps acceleration to the expected estimation error. This function serves as the foundation for our slip-based control algorithm design, which will be further discussed in the experimental section.

In practice, each spatial location may typically be sampled only once during the measurement process. However, by intentionally routing the autonomous vehicle (AV) through the same location multiple times, it is possible to obtain multiple independent observations. Let $\mathcal{K}$ denote the set of such observations collected at a specific location $L$. The aggregated peak TRFC estimate at location $\bar{y}^L$ is: 
\begin{align}
    VAR(\bar{y}^L)=(\sum_{k\in\mathcal{K}}{\frac{1}{\delta _k^2}})^{-1}
    \label{eq:variance_reduction}
\end{align}

Equation (\ref{eq:variance_reduction}) demonstrates that prediction error can be further minimized by aggregating multiple independent observations.

\subsection{High slip ratio controller design}
\label{sec:high_slip_controller}

\begin{figure}
    \centering
    \includegraphics[width=0.8\linewidth]{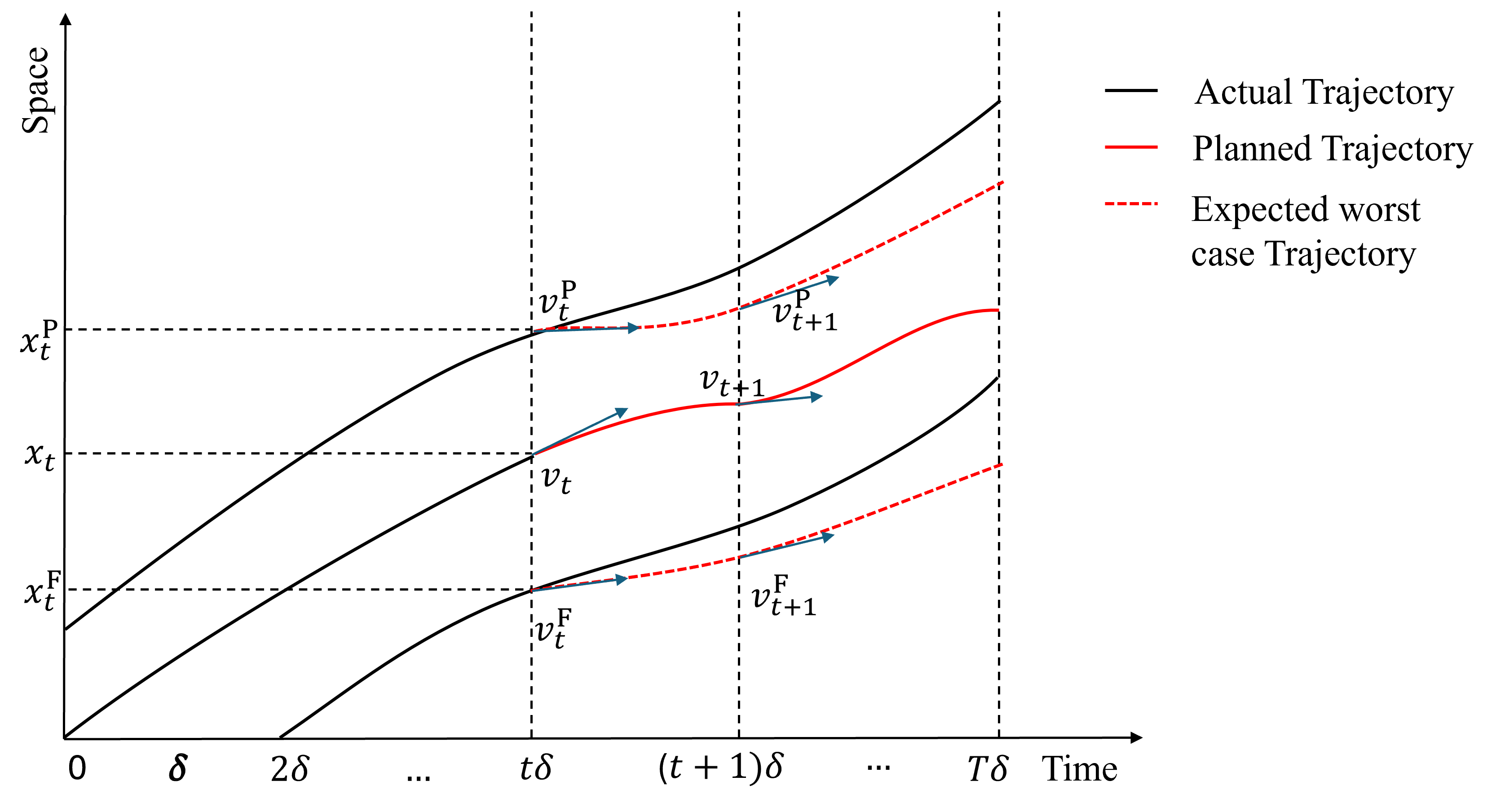}
    \caption{Illustration of controller status}
    \label{fig:Time-space}
\end{figure}

As illustrated in Figure \ref{fig:Time-space}, we investigate a scenario where the ego AV drives in a car-following model interacting with a preceding human vehicle and a following vehicle. 

Consider a set of discrete time indexes  $\mathcal{T} = \{0,1,2,...,T\}$. Define $x_t$, $v_t$, and $a_t$ as the location, velocity, and acceleration of the ego AV at time $t\delta$  with a unit time interval $\delta,\forall t\in \mathcal{T}$. Similarly, define  $x^\text{P}_t,x^\text{F}_t,v^\text{P}_t,v^\text{F}_t$, and $a^\text{P}_t,a^\text{F}_t$ as the position, velocity, and acceleration of the preceding vehicle and following vehicle at time $\delta t$, respectively. Where the supperscript $\text{P}$ and $\text{F}$ refer to the preceding and following vehicles. Let $\lambda > 0$ be a regularization weight rewards control oscillations. 

To ensure the safety of the AV and surrounding vehicles during high slip ratio control, a worst-case safety assumption is adopted. Specifically, it is assumed that the preceding vehicle performs maximum deceleration while the induced acceleration of the following vehicle remains in the safety region. Under these conditions, the AV’s planned motion must guarantee that no collision occurs.
The assumptions are as follows:
\begin{itemize}
    \item Preceding vehicle maximum deceleration $-b$.
    \item Following vehicle follows an IDM model bounded by maximum allowed deceleration $a$.
    \item Ego AV's acceleration: $a_{\min } \leq a_{t} \leq a_{\max }$.
    \item The following optimization is only solved at time step where: $v_t>v_{threshold}=2m/s$.
\end{itemize}

The high slip ratio controller can then be formulated as the following optimization problem:
\begin{align}
    \min \quad & \sum_{t\in \mathcal{T}}[e(a_t) - \lambda\cdot (a_t - a_{t-1})^2] \\
    \text{s.t.}  \quad  & v^\text{P}_{t+1} =\max\{ v^\text{P}_t-b\delta,0\}, & \forall t\in \mathcal{T} \backslash \{T\},\\
    &x^\text{P}_{t+1} =x^\text{P}_t + (v^\text{P}_t+v^\text{P}_{t+1})\cdot\delta/2, & \forall t\in \mathcal{T} \backslash \{T\},\\
    &v_{t+1} = \max\{v_t + a_t \cdot \delta,0\}, & \forall t\in \mathcal{T} \backslash \{T\}, \\
    &x_{t+1} = x_t + v_t \cdot \delta  +\frac{1}{2}a_t\cdot \delta ^2, & \forall t\in \mathcal{T} \backslash \{T\},\\
    &a_{\min } \leq a_{t} \leq a_{\max }, & \forall t\in \mathcal{T},\\
    &a^\text{F}_t = a_{IDM}(v^\text{F}_t,x^\text{F}_t,v_t,x_t)\leq a, & \forall t\in \mathcal{T},\\
    &v^\text{F}_{t+1} =\max\{ v^\text{F}_t-a\delta,0\}, & \forall t\in \mathcal{T} \backslash \{T\},\\
    &x^\text{F}_{t+1} =x^\text{F}_t + (v^\text{F}_t+v^\text{F}_{t+1})\cdot\delta/2, & \forall t\in \mathcal{T} \backslash \{T\},\\
    & x^\text{P}_t > x_t > x^\text{F}_t,  & \forall t \in \mathcal{T}.
\end{align}

    \section{Experiment}
\label{sec:exp}

To validate the effectiveness of the proposed high-slip-ratio control and TRFC estimation framework, we conduct experiments in both simulation and real-world settings. A Gaussian model is fitted offline to relate acceleration magnitude to estimation error (i.e. $e(a_t)$), enabling practical controller implementation. The objective is to assess whether the framework can (i) reliably excite the high-slip regime under realistic car-following constraints, (ii) accurately estimate the peak TRFC using projection-based statistical methods, and (iii) maintain safety and control feasibility throughout the maneuver.  

\subsection{Simulation Setup}

We implement the proposed high-slip-ratio controller and projection-based TRFC estimator in a closed-loop simulation environment developed in MATLAB/Simulink. The ego vehicle follows a preceding vehicle that applies maximum deceleration to emulate a worst-case car-following scenario. The control strategy is based on the optimal control formulation described in Section~\ref{sec:high_slip_controller}. Safety constraints are enforced throughout the simulation to ensure no collisions occur with either the preceding or following vehicle during high-slip excitation.

\subsection{Simulation Results and Analysis}

\begin{figure}
    \centering
    \includegraphics[width=1\linewidth]{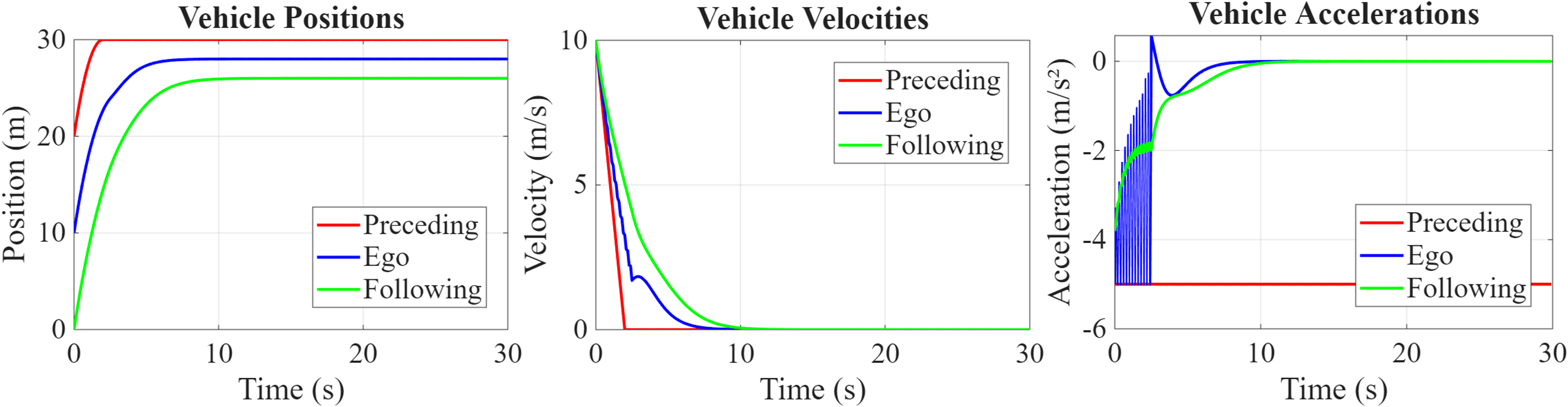}
    \caption{Simulation of controller in worst-case scenarios}
    \label{fig:simulation}
\end{figure}

The simulation results presented in Figure~\ref{fig:simulation} demonstrate the effectiveness of the proposed high-slip-ratio control framework in safely exciting the critical region of the tire–road friction curve. This excitation is sufficient to expose the peak TRFC, which is essential for reliable friction estimation. 

In the worst-case scenario where the preceding vehicle applies maximum deceleration, the ego vehicle successfully maintains safe longitudinal spacing from both the preceding and following vehicles. As shown in the position and velocity plots, all three vehicles exhibit smooth and stable trajectories, with the ego vehicle responding appropriately to the deceleration event without compromising safety. The acceleration profile of the ego vehicle shows initial oscillations during the slip excitation phase but quickly stabilizes within safe bounds. This controlled behavior indicates that the proposed controller effectively balances slip-ratio excitation with inter-vehicle safety constraints. The following vehicle also maintains sufficient headway, confirming that the overall system remains stable throughout the maneuver. 

These results validate a core contribution of this work: accurate and scalable TRFC estimation can be achieved through deliberate slip excitation using automated vehicles operating in realistic car-following scenarios. Importantly, the approach does not rely on any dedicated friction-sensing hardware, enabling practical deployment in real-world AV systems.

\subsection{Real-World Validation}

\begin{figure}
    \centering
    \includegraphics[width=0.7\linewidth]{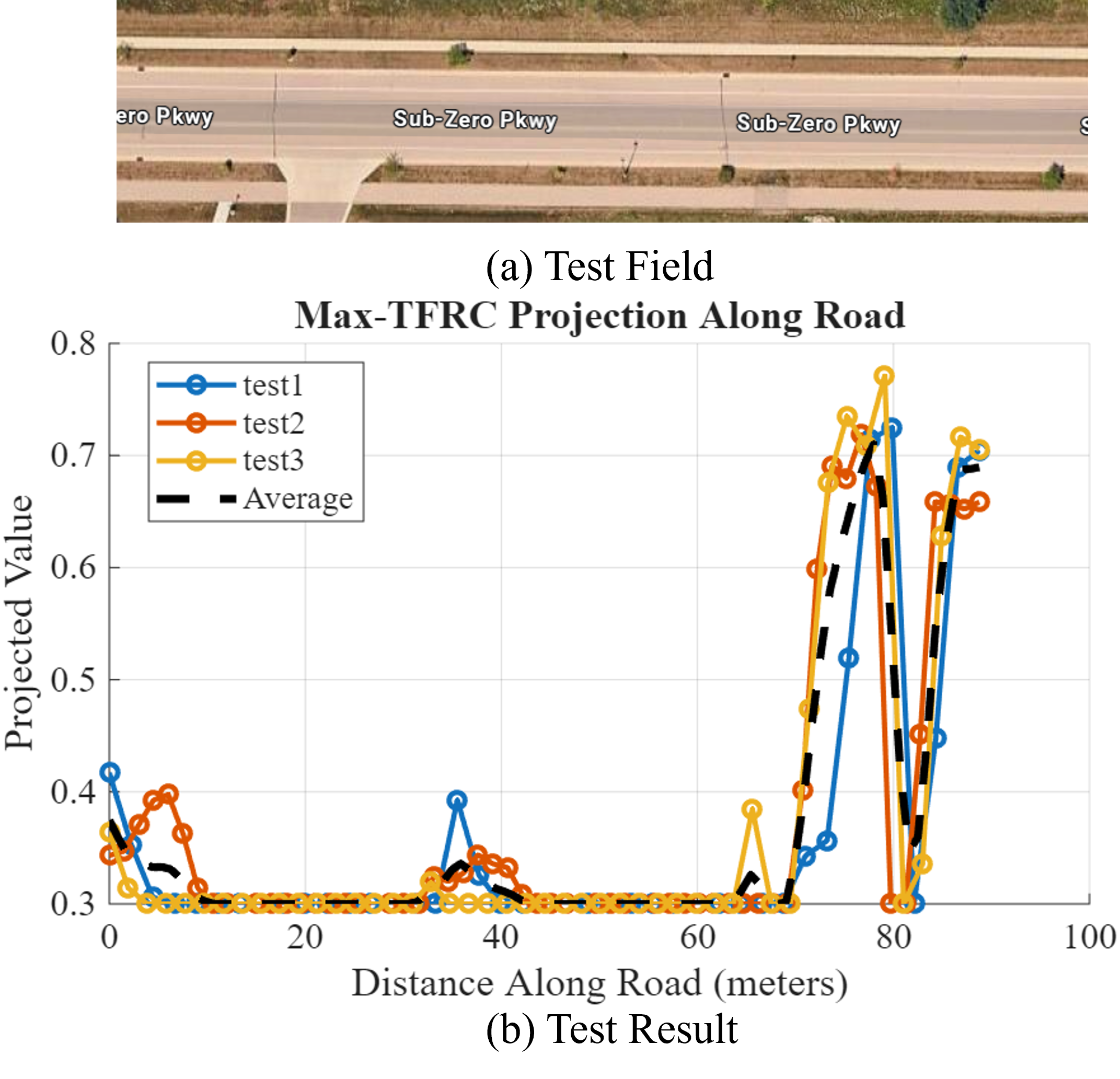}
    \caption{Field testing location (a) and result (b).}
    \label{fig:field}
\end{figure}

To evaluate the performance of the proposed friction estimation method under realistic operational conditions, we conducted a series of field experiments using our autonomous vehicle (AV) testbed. The AV platform is equipped with a high-resolution LiDAR, a GPS/INS unit for precise localization, and a control-by-wire interface that enables direct actuation of throttle, brake, and steering systems. The vehicle also supports real-time data logging and closed-loop execution of planned trajectories and acceleration profiles, making it well-suited for tasks involving longitudinal slip excitation and TRFC estimation.

The field tests were conducted along a designated stretch of road under dry concrete surface conditions (Figure \ref{fig:field}(a)). During each test, the AV executed a controlled sequence of acceleration and deceleration maneuvers, following a virtual lead vehicle trajectory while adhering to safety constraints defined by a car-following policy. This ensured repeatable slip generation while preserving safe longitudinal gaps.

TRFC values were estimated offline from the recorded sensor data using the proposed Simplified Magic Formula-based estimator. To improve robustness and mitigate local estimation noise, each test scenario was repeated multiple times. The resulting friction estimates were aggregated spatially using the variance-reduction approach introduced in Equation (\ref{eq:variance_reduction}).

As illustrated in Figure~\ref{fig:field}(b), the proposed estimation method produced consistent and spatially coherent TRFC projections across multiple independent test runs. The observed repeatability highlights the method’s robustness to environmental variability and sensor noise, supporting its potential for real-world friction mapping applications. It is important to note, however, that the reliability of the projection is inherently influenced by the vehicle’s executed maneuvers—smaller acceleration or deceleration inputs can result in increased estimation variance due to insufficient slip excitation. While the current setup, involving a single AV performing repeated test runs, provides a useful baseline for validating the projection approach, the methodology could be further enhanced by deploying a fleet of AVs executing this algorithm during routine empty-haul operations. Such a deployment would enable broader spatial coverage and statistical reinforcement of localized friction estimates. 
	\section{Summary and Conclusions}
	\label{sec:conclusion}
	This paper presents a high-slip-ratio active control and estimation framework to overcome the limitations of conventional TRFC measurement methods under low-excitation conditions. By leveraging the flexibility of AVs during empty-haul trips, the proposed method enables safe and deliberate excitation of high slip ratios, facilitating reliable observation of peak tire–road friction without requiring specialized hardware. A simplified Magic Formula tire model is adopted to capture nonlinear slip–force dynamics, while a binning-based statistical projection method improves robustness under local data sparsity and sensor noise. The high-slip-ratio controller is formulated as a constrained optimization problem that ensures excitation effectiveness, trajectory tracking, and inter-vehicle safety in car-following scenarios. The framework is validated through both closed-loop simulations and real-world AV experiments, demonstrating its accuracy, safety, and scalability for friction screening applications. Future work will focus on real-time implementation across diverse road surfaces, long-term deployment under varying environmental conditions, and integration with cooperative fleet-based friction sensing systems for infrastructure-level friction mapping.
	
	\section*{ACKNOWLEDGMENTS}
	This work was sponsored by the Center for Connected and Automated Transportation (CCAT) project “Roadway Friction Screening and Measurement with Automated Vehicle Telematics and Control” and by the National Science Foundation (NSF) under Grant No. 2343167.
	
	\bibliographystyle{IEEEtran}
	\bibliography{root} 
	
\end{document}